\documentclass{WileyMSP-template}

\usepackage{graphicx}
\usepackage{epstopdf}
\usepackage{ragged2e}
\usepackage{amsmath,amsfonts,amssymb,amsthm}
\usepackage{cite}
\usepackage{float}

\usepackage{xcolor}
\usepackage[font=sf,labelfont=bf]{caption}

\begin{document}

\pagestyle{fancy}
\rhead{\includegraphics[width=2.5cm]{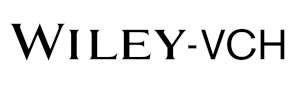}}

\title{Self-Sustained And Coordinated Rhythmic Deformations With SMA For Controller-Free Locomotion}
\maketitle

\author{Ziyang Zhou* and Suyi Li}

\begin{affiliations}

Department of Mechanical Engineering, Virginia Tech\\
181 Durham Hall, 1145 Perry Street, Blacksburg, VA 24061, USA\\[0.1in]

*Correspondent email address: \texttt{zzhou4@vt.edu}

\end{affiliations}


\begin{abstract}
\justifying
\it{
This study presents a modular, electronics-free, and fully onboard control and actuation approach for SMA-based soft robots to achieve locomotion tasks. This approach exploits the nonlinear mechanics of compliant curved beams and carefully designed mechanical control circuits to create and synchronize rhythmic deformation cycles, mimicking the central pattern generators (CPG) prevalent in animal locomotions. More specifically, the study elucidates a new strategy to amplify the actuation performance of the shape memory coil actuator by coupling it to a carefully designed, mono-stable curve beam with a snap-through buckling behavior. Such SMA-curved beam assembly is integrated with an entirely mechanical circuit featuring a slider mechanism. This circuit can automatically cut off and supply current to the SMA according to its deformation status, generating a self-sustained rhythmic deformation cycle using a simple DC power supply. Finally, this study presents a new strategy to coordinate (synchronize) two rhythmic deformation cycles from two robotic modules to achieve efficient crawling locomotion but still use a single DC power. This work represents a significant step towards fully autonomous, electronics-free SMA-based locomotion robots with fully onboard actuation and control.
}
\end{abstract}

\section{Introduction}

\justifying

The recent advances in bio-mimicry, material science, design methodology, and control theory have enabled us to build genuinely soft robots that can collaborate closely with humans in unstructured and dynamic working environments \textcolor{red}{\cite{gupta2019soft}}. These robots are typically made of soft materials exhibiting low elastic modulus and high strain before failure, making them safer and more versatile than the traditional rigid-linked robots for tasks like surgery \textcolor{red}{\cite{cianchetti2018biomedical}}, rehabilitation \textcolor{red}{\cite{polygerinos2015soft}}, search/rescue \textcolor{red}{\cite{stokes2014hybrid}}, and even space operations \textcolor{red}{\cite{zhang2023progress}}. 
However, despite these promising potentials, many challenges remain unsolved due to the complex and nonlinear nature of the soft robotic body and the associated actuation apparatus \textcolor{red}{\cite{rus2015design}}.

\medskip
Among these challenges, one of the most significant is devising \textit{onboard} actuation and control. 
Many actuation mechanisms have been examined in soft robotic systems with success, including magnetic field \textcolor{red}{\cite{xiong2023fast}}, pneumatic pressure \textcolor{red}{\cite{decker2022programmable, drotman2021electronics, zhai2023desktop, wang2023magnetic}}, adaptive structures \textcolor{red}{\cite{wang2023building, deshpande2023high}} and responsive materials \textcolor{red}{\cite{wang2023versatile, he2023modular}}. Their corresponding control units, however, are usually ``off-board'' and cumbersome (e.g., pneumatic transducers connected to computers). This prevents the soft robots from being deployed fully autonomously. Therefore, there is a need to integrate both actuation and its control components onboard the compliant robotic body.

\medskip
To address this challenge, researchers have explored integrating small-scale electronic control boards with piezoelectric actuators to achieve precisely controlled and fast locomotion \textcolor{red}{\cite{miao2023power}}. While piezoelectric actuators offer a high-frequency response \textcolor{red}{\cite{chen2023bio}} and can be powered by a compact energy supply \textcolor{red}{\cite{ma2023compact}}, they can only generate a small strain, making them less suitable for large-scale robots. More importantly, using rigid electronic control boards is not ideal for soft robots. Alternatively, there is a recent interest in developing entirely pneumatic logic circuits and actuators to execute control commands \textcolor{red}{\cite{decker2022programmable}}, complex locomotion, self-sensing \textcolor{red}{\cite{drotman2021electronics}}, and accurate gripping \textcolor{red}{\cite{zhai2023desktop}}. Such an integrated pneumatic control and actuation setup is electronics-free and can be seamlessly embedded into a soft robotic body. However, it often requires a bulky pressure supply, and its design can be pretty complicated, making it challenging for miniaturization and leak-proofing.
\medskip

Besides piezoelectric and pneumatic actuators mentioned above (and many more exotic active materials), SMA shape memory alloy actuators (or similar shape memory polymers) have been the preferred choice for many soft robotic systems because they are robust, inexpensive, and capable of providing a large actuation force with a substantial stroke. Nonetheless, SMA actuators are sensitive to ambient temperature and exhibit nonlinear mechanical responses, so effective control is crucial. To this end, one can use an electronic control unit off-board \textcolor{red}{\cite{firouzeh2020stretchable, huang2019highly}}, but there is a potential to exploit the nonlinear mechanics of compliant structures to simplify the control process and enable an onboard setup. For example, integrating a bi-stable beam with multiple shape memory actuators can generate crawling locomotion like the earthworm \textcolor{red}{\cite{baek2023sma, patel2023highly}} or turtle \textcolor{red}{\cite{baines2021amphibious, baines2019toward}}, even creating mechanical sensing and logic gait operation \textcolor{red}{\cite{yan2023origami}}. Nonetheless, many untapped potentials remain to be explored by integrating mechanical components with SMA actuators to enable simple onboard actuation and control.

\begin{figure}[t!]
    \centering
    \includegraphics[scale=1.0]{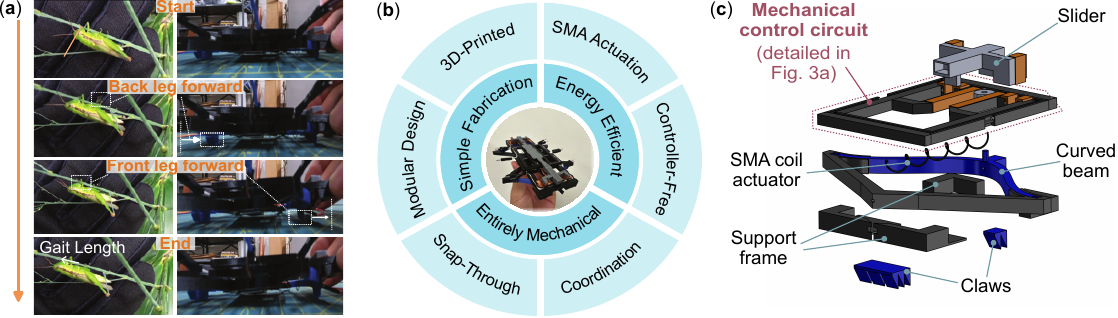}
    \caption{\textbf{The overall concept of the modular robots in this study.} (\textbf{a}) A comparison between the walking pattern of the Chinese grasshopper and our two-module robot reveals many similarities in their gait coordination. (\textbf{b}) Key features of the modular robot platform.(\textbf{c}) An exploded view of a single robotic module; key components are labeled, and each module is 66 mm in length. }
    \label{fig:bigpic}
\end{figure}

\medskip
This study introduces an SMA-based and modular robot design with integrated onboard actuation and control to achieve crawling locomotion without any electronics. This design took inspiration from the central pattern generators (CPG) that are prevalent in animal locomotion. In the CPG framework, the biological neural circuits produce rhythmic motor patterns without sensory or descending inputs (e.g., from the brain) that carry specific timing information. Several rhythmic patterns can then coordinate to generate complex locomotions, such as walking, undulating, flying, and swimming \textcolor{red}{\cite{li2021self, tolley2014untethered,shepherd2011multigait,hwang2022shape}}. Similarly, each module in our robot can generate a self-sustained rhythmic deformation with a constant electric current input, and two identical modules can coordinate their deformations (also entirely mechanically) to produce crawling locomotion.  Such crawling locomotion shares a lot of similarities with the crawling gait of the Chinese hopper on a thin branch (Figure \ref{fig:bigpic}a,b).

\medskip
To achieve the CPG-like rhythmic mechanical deformation, each module in our robotic system incorporates two unique features:  A \textit{mono-stable}, curved beam exhibiting a nonlinear snap-through force-deformation relationship. This beam is connected to a \textit{single} SMA actuator as a ``bias spring'' to amplify its stroke. The second unique feature is a mechanical slider that automatically disconnects and reconnects the SMA from the electric current supply, producing self-sustained, rhythmic mechanical deformations. The robotic module is 3D printed on a commercial printer, ensuring low cost and a simple fabrication process (Figure \ref{fig:bigpic}c). Moreover, to combine and coordinate two rhythmic patterns from two modules, we designed a mechanical ``control circuit'' involving a compliant, curved bistable beam and a slider. This mechanism enables the two robotic modules to synchronize their deformation with only a single constant current supply, producing a more efficient and faster crawling gait.

\medskip
Overall, this study expands the possibilities of using SMA to create entirely onboard actuation and control for soft robots and demonstrate an efficient and versatile modular design. The results can be extended for many locomotion tasks beyond crawling. 

\section{Result}

\subsection{Actuation Mechanism Design}
Figure \ref{fig:bigpic}(c) illustrates the overall design of a single robotic module, which consists of six components. They are, from top to bottom: (1) a 3D-printed slider with the top of its branches wrapped with cooper sheets for electric conductivity; (2) a ``mechanical control circuit'' (further discussed in section 2.2); (3) a shape memory alloy (SMA) coil serving as the primary actuation source; (4) a mono-stable curved beam 3D-printed with TPU 95A and nylon, (5) a 3D-printed support frame, and (6) a pair of claws producing anisotropic friction. The curved beam and SMA coil are the most critical components here because their mechanical properties directly dictate the generation of self-sustained rhythmic deformation and overall locomotion performance.

\subsubsection{Selecting the SMA Coil Actuator}
\begin{figure}[hbt!]
    \centering
    \includegraphics[scale=1.0]{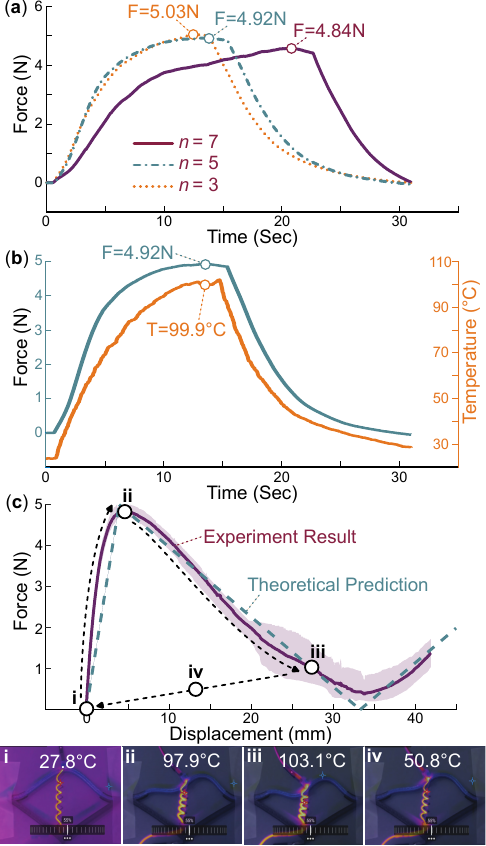}
    \caption{\textbf{Design, test, and integration of SMA coil actuator and curved snap-through beam.} (\textbf{a}): A comparative study of SMA actuators with different coil numbers ($n$), showing the relationship between heating time and output force. (\textbf{b}) Output force and temperature correlation of the $n=5$ SMA coil. (\textbf{c}) The theoretically predicted and measured force-deformation curve of the curved beam.(The solid line is the averaged test data from 4 load cycles, and the shaded region is the standard deviation). The infrared images below show the deformation of the SMA-curved beam assembly at different deformation stages.}
    \label{fig:SMABeam}
\end{figure}

To find the most suitable SMA coil actuator for modular robots, one must select several design parameters, including the transition temperature, mandrel size, and number of coils ($n$). To this end, we tested the actuation performance of several SMA coils with different transition temperatures and mandrel sizes (Kellogg's Research Labs, details in supplementary materials section 1). We chose a mandrel size of 4.75mm, a wire diameter of 0.5mm, and a transition temperature of 45℃. Another critical parameter is the number of SMA coils, which not only constrains the overall size of the robotic module but also determines the SMA's output force. Test results shown in Figure \ref{fig:SMABeam}(a) indicate that the SMA's output force increases with a smaller coil number. However, the shortest SMA coil may not match the curved beam's mechanical response (more on this later). It is worth noting that SMA coils with $n=3$ and $n=5$ exhibit a similar output force and heat-up time, but the 5-coil one is longer, offering a more substantial locomotion stroke. Therefore, we subsequently focus on the SMA coil with $n=5$. 
\medskip

SMA is sensitive to temperature, so it is crucial to examine the correlation between its output force and temperature. The test results in Figure \ref{fig:SMABeam}(b)  depict the relationship between the SMA coil's output force, temperature, and time using Joule heating (3V DC power supply, TENMA 72-6854). It is worth noting that the SMA coil must be heated to its maximum operational temperature to achieve its peak force, and it takes approximately 14 seconds to reach this peak from room temperature. However, later in the locomotion tests involving repeated cycles of heating and cooling SMA, the heating time in each cycle will be shorter since the SMA has already been pre-heated. These test results of SMA's output directly inform the design of the mono-stable curved beam and its snap-through response, as we detail in the next section.  

\subsubsection{Designing the Curved Beam and its Snap-Through Response}
The curved beam consists of two 3D-printed parts: the beam body made of TPU 95A material, and a supporting frame made of Nylon. They are printed as one piece using a dual-nozzle FDM printer (Ultimaker S5). The design of the supporting frame draws inspiration from an ancient weapon, the crossbow. It has two purposes: enhancing the overall structural stiffness. 
This curved beam has a non-monotonic, snap-through force-deformation relationship. Its reaction force will first rise when transversely loaded at its center. However, when the external force reaches a critical level, elastic instability occurs. As a result, the curved beam shows a negative stiffness over an extensive deformation range (with a decreasing reaction force), resulting in a ``snap-though'' behavior as the beam quickly deforms into an inverted curve shape. Eventually, the reaction force rises again at a much higher deformation.  

\medskip
Using this curved beam as the ``bias spring'' for the SMA coil actuator has a significant advantage over the traditional linear spring, and this is one of the major contributions of this study. If the heated SMA actuator can generate a force that matches the curved beam's critical load, it can trigger the snap-through response, resulting in a large and rapid actuation stroke. Then, when the SMA starts cooling down, the mono-stable curved beam can provide a relatively large reaction force over a significant deformation range to quickly return the SMA to its original configuration (Figure \ref{fig:SMABeam}c, supplementary movie S1). 

\medskip
Based on this working principle, we used a theoretical model \textcolor{red}{\cite{hussein2015modeling}} to design the snap-through beam (details in Section 2 of supplementary materials). The final design is expected to exhibit a theoretical snap-through force of 4.92N, which precisely matches the peak output force of the selected SMA coil. The experimental measured snap-through force, as shown in Figure \ref{fig:SMABeam}(c), is $4.82 \pm 0.1468$N. The minor discrepancy probably originates from the nature of 3D printing. For example, the soft TPU 95 material's elastic modulus always differs from the theoretical value due to different printing setups \textcolor{red}{\cite{rodriguez2021influence}}. Regardless, the experimentally measured force-deformation curves are consistent throughout repeated loading cycles, ensuring a consistent rhythmic deformation (more in the following sections). 

\subsubsection{Integrating SMA and Curved Beam}
After the comprehensive analysis and design of the SMA coil and curved beam, we assembled them to verify their integration. Figure \ref{fig:SMABeam}(c) shows a complete actuation cycle and the infrared images of the SMA-curved beam assembly (FOTRIC 346A w/25$^{\circ}$ Lens). The results confirm that the SMA coil reaches its desired temperature at the curved beam's snap-through point, indicating that the SMA is outputting its maximum force. Subsequently, with an additional 15 seconds of heating, the curved beam reaches its target snap-through deformation. In this test, the power supply to the SMA is manually disconnected to initiate the cooling process (supplement video S1). 

\medskip
It is worth noting that in this verification test, the curved beam exhibits the second snap-through mode in that it deforms into a more complex shape resembling a complete sine wave. This phenomenon is common for a slender structure \textcolor{red}{\cite{kapania2003formulation,vo2020total}}. Regardless, the SMA-curved beam assembly achieved the desired deformation of 25mm, suggesting that the snap-through force in the second mode is close to the theoretical calculation. 

\subsection{Rhythmic Deformation from a Single Module}
\begin{figure}[hbt!]
    \centering
    \includegraphics[]{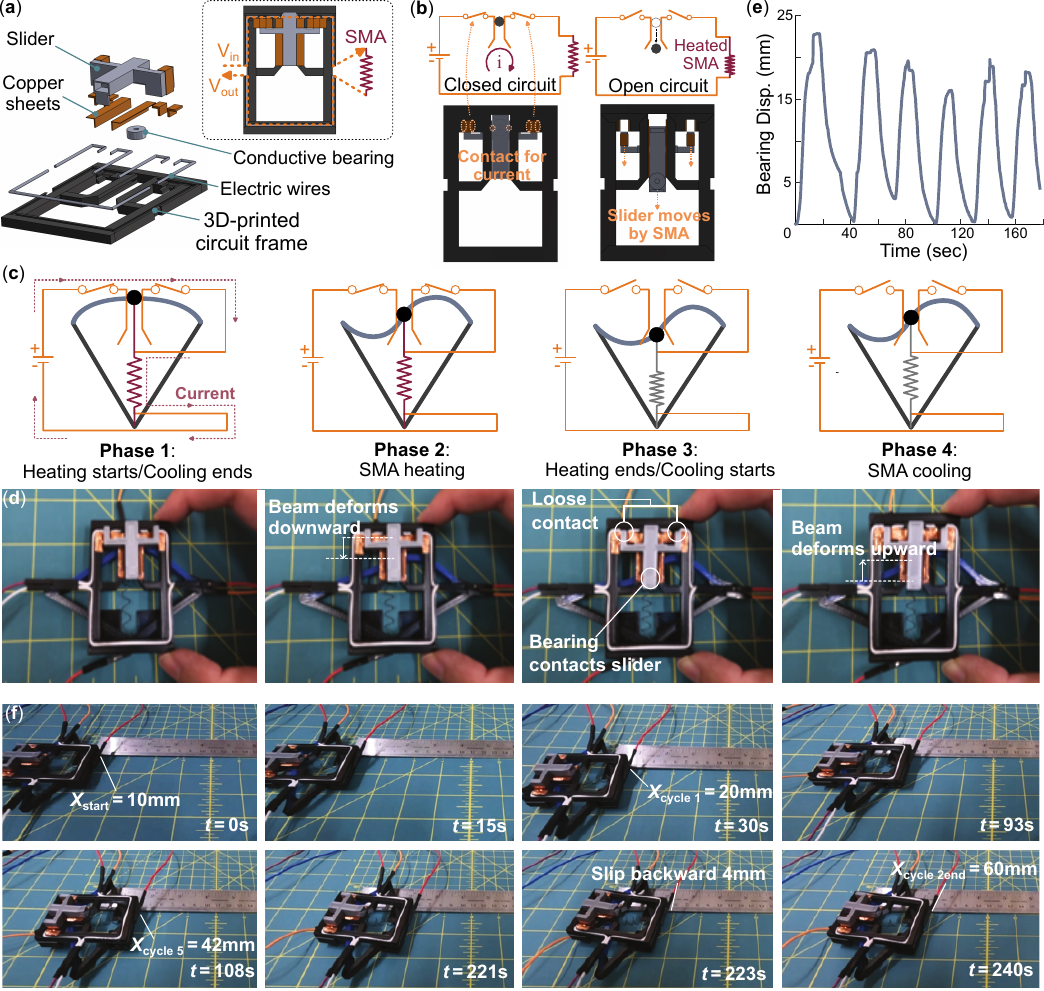}
    \caption{\textbf{The working principle of the rhythmic deformation from on a single robotic module and the corresponding crawling locomotion} (\textbf{a}) Exploded view of the ``mechanical control circuit'' (Fig. \ref{fig:bigpic}c). (\textbf{b}) The mechanical slider can move out of (or into) the circuit frame's slots, thus stopping (or starting) the electric current. (\textbf{c ,d}) The schematic diagram and experiment photos show the four phases of a rhythmic deformation cycle. (\textbf{e}) The experimentally measured rhythmic deformation, the inconsistency between different cycles is probably due to the unpredictable sliding friction contact between the slider and circuit frame. (\textbf{f}) A single module, equipped with 3D-printed claws, completed 12 deformation cycles and traveled 50mm, all using a simple DC power supply.}
    \label{fig:onemodule}
\end{figure}

The ``mechanical control circuit'' shown in Figures \ref{fig:bigpic}(c) and \ref{fig:onemodule} plays a vital role in generating the CPG-like rhythmic deformation and locomotion gait. As a result, this modular robot introduced in this paper stands out from others in that it can accomplish locomotion tasks even with only one module. This mechanical control circuit consists of four components (Figure \ref{fig:onemodule}a): (1) a 3D-printed ``circuit frame'' that connects the curved beam, SMA coil, and other components; (2) embedded electric wires in the circuit frame; (3) a conductive bearing attached at the center of the curved beam; (4) a mechanical slider to enable the automatic start and stop of electric current.
On the top side of the circuit frame, there are two slots wrapped with copper sheets for the sliders to start and stop the current flow. At the circuit frame's center, there is a sliding rail also covered with copper sheets to accommodate the conductive bearing. These copper sheets guarantee the passage of electric current and smooth bearing movement. The mechanical slider features two stoppers at the top and bottom of its central portion, set at a distance of 25mm from each other, precisely matching the desired travel distance of the curved beam (Figure \ref{fig:onemodule}b).

\subsubsection{Working Principle of a Single Module}
When DC power is supplied to the robotic module, all its components will cooperate to generate rhythmic mechanical deformation cycles, and each cycle can be divided into four phases (Figure \ref{fig:onemodule}c,d). 
In the first phase, the curved beam is in its initial shape, the mechanical slider sits in the circuit frame's slots, and the conductive bearing is within the sliding rail (aka, the circuit is closed). Therefore, during this first phase, the electric current can flow through the SMA coil, quickly heating it up. 
In the second phase, the SMA's temperature is high enough to produce its peak force and trigger the snap-through deformation of the curved beam. As a result, the conductive bearing attached at the center of the curved beam moves along the sliding rail. However, the mechanical slider remains at its original position. As a result, the overall electric circuit is still closed, and SMA heating continues. 
When the SMA-curved beam reaches its targeted deformation of 25mm, the conductive bearing touches the bottom stopper on the mechanical slider. This marks the transition into the third phase. In this phase, the conductive bearing pushes the bottom stopper until it moves the mechanical slider out of the circuit frame's slot, cutting off the electric current flow. As a result, the SMA coil starts to cool down.
In the final fourth phase, the cooling SMA reverts to its original length due to the nonlinear restoring force from the curved beam. As a result, the conductive bearing moves up along the circuit frame's sliding rail. However, the mechanical slider remains disengaged with no electric current. Such SMA cooling and restoration continues until the curved beam stretches the SMA back to its original length. At this moment, the conductive bearing touches the upper stopper on the mechanical slider, pushing it back into the circuit frame's slot. As a result, the robotic module reset itself to the original configuration (except for some residual heat in the SMA coil), the electric current starts flowing again, and the next deformation cycle begins (supplement video S2).

\medskip
Figure \ref{fig:onemodule}(e) shows the experimentally measured rhythmic deformation cycles from the single module. It is worth noting that the reason for having two branches on the mechanical slider (and two slots in the circuit frame) instead of one is to ensure the slider can disconnect the electric current reliably at the end of phase 2. There is a slight possibility for the slider to get stuck in the slot while the bearing pushes the bottom stopper, mainly due to a misalignment of the copper sheet. A second slot/branch can significantly reduce such failure (supplement video S2). 

\subsubsection{Single-Module Robotic Locomotion}

We attach 3D-printed ``claws'' to the robotic module and demonstrate its locomotion capability by connecting it to a DC power supply that provides a constant 3.5V and 2.5A electricity. The robot spent 240 seconds to complete 12 full deformation cycles and traveled 50mm, and 20 seconds per cycle (supplement video S3). Note that the first cycle took the longest to complete because the SMA coil starts from room temperature, requiring a longer heating to reach its peak temperature. However, after the first cycle, SMA has some residual heat, so subsequent cycles need less heating, resulting in shorter actuation cycles. 

\medskip
One of the primary challenges affecting the robot’s speed is the backward slip. During the first phase of each actuation cycle, the robot slipped backward. One can undoubtedly address this issue with a better surface contact design \textcolor{red}{\cite{blau2001significance,filippov2013frictional}}. However, this is beyond the scope of this study.

\subsection{Synchronizing Rhythmic Deformations from Two Modules}
To further mimic the CPG-based locomotions in the animal kingdom, we integrate two identical robotic modules and devise an entirely mechanical approach to coordinate (synchronize) their rhythmic deformations. The dual modular robot has five components: two robotic modules, a ``module connector'' to ensure a seamless corporation between these two modules, and a newly designed slider mechanism (Figure \ref{fig:twomodule}a). It is this new slider mechanism and module connector that can achieve the CPG-like locomotion gait coordination, so we detail their design and working principles in the following.

\subsubsection{Bistable Switch and New Slider Design}
The module connector comprises a 3D-printed circuit frame, conductive wire, copper sheet, and a curved bistable beam (referred to as a ``bistable switch'' to avoid confusion with the curved monostable beam within a single robotic module). The bistable switch is also 3D-printed with TPU 95A material, featuring a double-curve design to prevent the second mode from occurring during the snap-through. This design ensures reliable switching when desired. The bistable switch has a critical snap-through force of 0.45N, as determined by the theoretical method \textcolor{red}{\cite{yan2023development}} (details in the supplementary materials section 3). Finally, copper sheets are wrapped around the top and bottom portions of the bistable switch and four connection points (aka terminals) on the circuit frame, providing a consistent electric flow. 

\medskip
The newly designed slider mechanism is longer and spans the two robotic modules. The two ends of this slider still feature the two stoppers (25mm apart) that work with the open slots and conductive bearing in each robotic module. In addition, the middle portion of this slider has two additional stoppers that work with the bistable switch. The distance between these two additional stoppers is one and a half times the length of the bistable switch. his particular specification has been derived through a series of experimental trials. As a result, when the mechanical slider moves, it can push and pull the bistable switch between its two stable states.

\begin{figure}[hbt!]
    \centering
    \includegraphics[]{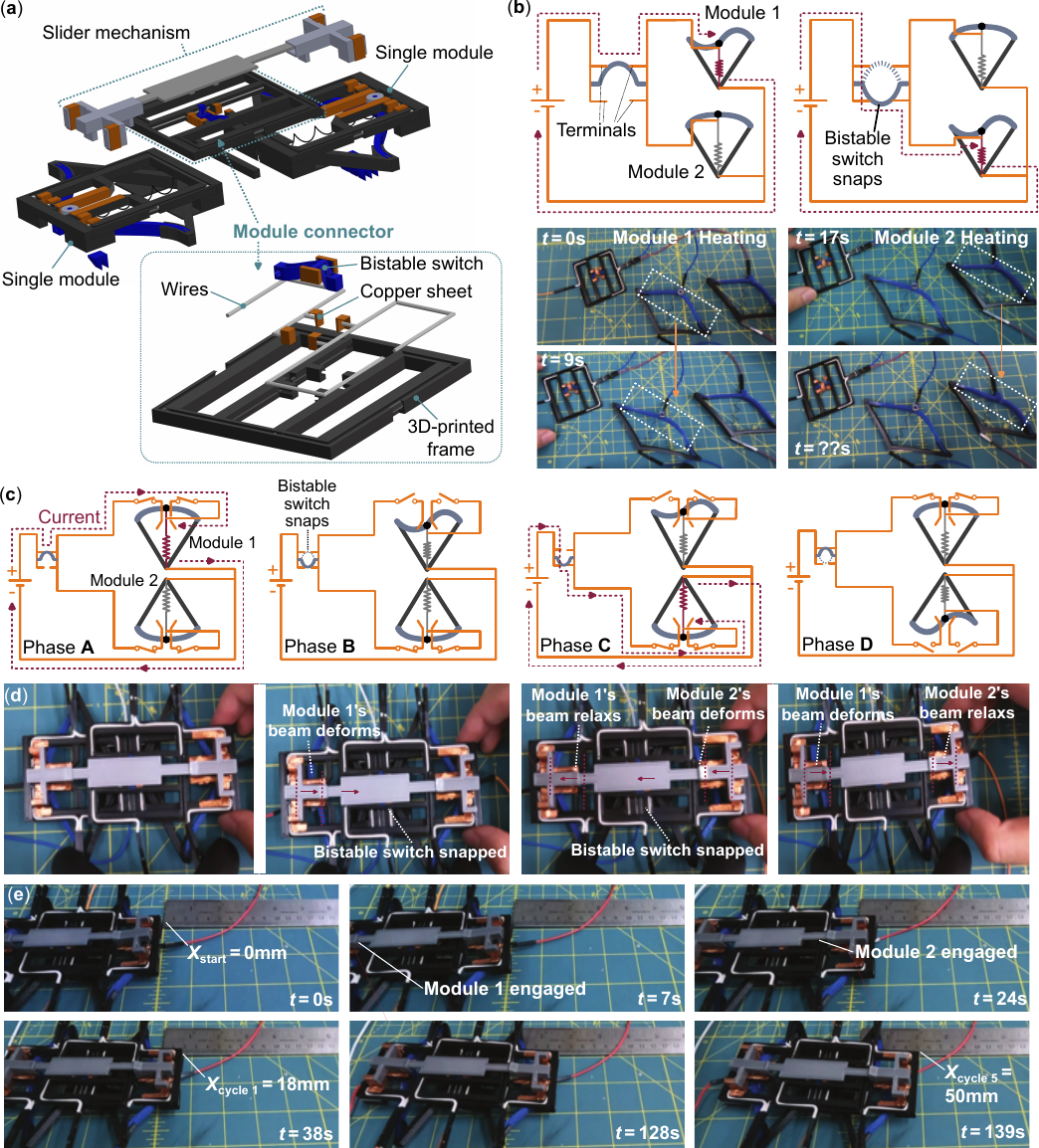}
    \caption{\textbf{Rhythmic deformation coordination with two robotic modules and improved crawling locomotion.} (\textbf{a}) Exploded View of a dual-module robot, highlighting the module connector's design. (\textbf{b}) Schematic diagram and experiment video to show the working principle of the module connector. (\textbf{c,d}) Overall working principle of the dual-module robot and its four phases of deformation synchronization: Schematic diagrams are at the top, and experiment photos are below. (\textbf{e}) The dual-module robot completes 5 coordinated deformation cycles and travels 50mm.}
    \label{fig:twomodule}
\end{figure}

\medskip
Implementing the bistable switch and newly designed mechanical slider gives the module connector two operation modes with a single DC power source (Figure \ref{fig:twomodule}b). In the first mode, the bistable curve settles at its first stable state, allowing the current to pass through the upper terminal and power the first robotic module. In the second mode, the bistable switch snaps to its second stable state, and the electric current is redirected to the lower terminal, powering the second module. This design ensures that the power is provided to only one module at a time (supplement video S4). 

\medskip

\subsubsection{Working Principle of Modular Synchronization/Coordination}

\medskip
The combination of the new slider mechanism and the module connector can ``detect'' each module's configuration and direct electric current from a single power source to either module accordingly. 
As a result, the dual-module robot can synchronize its two rhythmic deformation cycles and complete a more elaborated actuation through four consecutive phases (Figure \ref{fig:twomodule}c, d). In the first Phase A, the bistable switch in the module connector is in its first stable state, allowing the electric current to power robotic module 1. As a result, only module 1’s SMA coil is heated and deforming its curved monostable beam. 
The second Phase B starts when module one’s curved beam reaches the desired deformation, causing the attached conductive bearing in the middle to contact the mechanical slider’s bottom stopper, pushing it downward. Subsequently, the slider moves its stopper in the middle and snaps the bistable switch into its second stable state. 
As a result, the electric current to module 1 is cut off and redirected to module 2. 
During the third Phase C, the module connector ``detects'' that module 1 has finished deforming because the bistable switch has snapped. The electric power now flows to module 2. As module 2's SMA coil starts heating and deforming, module 1's SMA begins its cooling process. 
The final Phase D occurs when the curved beam in module 2 achieves the desired deformation and pushes the slider, snapping the bistable switch back to its first stable state. During the last phase, module 1’s SMA coil has cooled down and is ready to be actuated again. 
These four phases work in a sequence, essentially synchronizing the rhythmic deformations of the two robotic modules so that they are always out of phase (supplement video S5).

\subsubsection{Two-module Robotic Locomotion}
Similar to the single-module robot discussed earlier, we attach 3D-printed claws to the dual-module robot to test its locomotion performance. Note that the two modules are assembled in opposite orientations, so we must orient the claws so that module 1 has low friction during its SMA coil’s heating and higher friction during cooling, but vice versa for module 2. As a result, the two-module robot achieves unidirectional crawling (besides some backward slippage, as we discussed earlier).

\medskip
In the locomotion tests, the dual-module robot still connects to a single DC power supply, providing a steady 3.5V and 2.5A of electric power. The robot traveled a distance of 50mm using five actuation cycles and 142 seconds in total (Supplement videos S6, S7). The average speed is 10mm per cycle, more than twice the speed of the single-module robot. Such improvement is due to the fact that, in the single-module robot, only half of the actuation cycle (aka when SMA is contracting) is used for active crawling. In the dual-module robot, the whole actuation cycle is used for crawling.  This clearly demonstrates the advantages of using rhythmic coordination for locomotion tasks.

\section{Discussion}
This study presents a modular, electronics-free, and fully onboard control and actuation approach for SMA-based soft robots to achieve locomotion tasks. This approach exploits the nonlinear mechanics of compliant curved beams and carefully designed mechanical control circuits to create and synchronize rhythmic deformation cycles, mimicking the central pattern generators (CPG) prevalent in animal locomotions. The most significant contributions of this study are three folds:

\medskip
First, we find a way to harness the non-monotonic force-deformation relationship of the mono-stable curved beam to amply the shape memory alloy (SMA) coil's actuation performance. We carefully design the curved beam's geometry so that its critical force at the snap-through buckling matches the peak actuation force of the SMA coil. In this way, we can use the curved beam as a nonlinear bias spring for SMA, producing a much more significant deformation than the traditional linear bias spring. 

\medskip
Secondly, we design a mechanical control circuit --- involving a slider mechanism with stoppers --- to automatically cut off or supply electric current to the SMA-curved beam assembly according to their deformation status. In this way, a robotic module can use a simple DC power supply to produce self-sustained rhythmic deformation cycles, which can be used to create crawling locomotion.

\medskip
Third, we devise another mechanical control circuit (aka. modular connector) to synchronize the rhythmic deformations of two robotic modules. This circuit combines a more elaborate slider mechanism and a bi-stable switch to power two modules alternatively, ensuring their deformation cycles are always in the opposite phase. In this way, a dual-module robot can produce faster and more efficient crawling locomotion (more than twice as fast as single-module locomotion) but still use a single DC power supply.

\medskip
Besides the electronics-free onboard control and actuation via CPG-like synchronized deformation cycles, the proposed robot is also lightweight and simple to fabricate, with most of its body 3D printed. Moreover, the modular setup allows for expanding current results to more elaborate locomotions, such as four coordinated modules for legged locomotion or even more modules for snake-like undulations. Overall, the results of this study open an avenue toward physically intelligent robots at the centimeter scale.

\medskip
\textbf{Supporting Information} \par
Supporting Information is available in the supplementary documents and videos.

\medskip
\textbf{Acknowledgements} \par 
The authors acknowledge the support from the National Science Foundation (CMMI-2239673) and Virginia Tech (via startup fund).

\medskip

%

\bibliographystyle{MSP}
\bibliography{reference}




\end{document}